\documentclass{article}

\PassOptionsToPackage{numbers}{natbib}

    \usepackage[preprint]{neurips_2022}

\usepackage[utf8]{inputenc}
\usepackage[T1]{fontenc} 
\usepackage{hyperref}
\usepackage{url}
\usepackage{booktabs}       
\usepackage{amsfonts}       
\usepackage{nicefrac}       
\usepackage{microtype}      
\usepackage{xcolor}         
\usepackage{amsmath}
\usepackage{caption}
\usepackage{subcaption}
\usepackage{graphicx}
\usepackage{wrapfig}
\usepackage[bottom]{footmisc}

\hypersetup{
     colorlinks   = true,
     citecolor    = gray,
     linkcolor   = [rgb]{0.4,0,0},
     urlcolor =  blue,
     filecolor = blue,
}

\title{Compete and Compose: Learning Independent Mechanisms for Modular World Models}

\author{
  Anson Lei\\
  Applied AI Lab\\
  University of Oxford, UK\\
  \texttt{anson@robots.ox.ac.uk} 
  \And
  Frederik Nolte\\
  Applied AI Lab\\
  University of Oxford, UK\\
  \texttt{frederiknolte@robots.ox.ac.uk}
   \And
   Bernhard Sch\"{o}lkopf \\
   MPI for Intelligent Systems\\
   T\"ubingen, Germany\\
   \texttt{bs@tue.mpg.de} 
   \AND
   Ingmar Posner \\
   Applied AI Lab \\
   University of Oxford, UK\\
   \texttt{ingmar@robots.ox.ac.uk} 
}

\begin{document}

\maketitle

\begin{abstract}
We present \textit{COmpetitive Mechanisms for Efficient Transfer} (COMET), a modular world model which leverages reusable, independent mechanisms across different environments. COMET is trained on multiple environments with varying dynamics via a two-step process: competition and composition. This enables the model to recognise and learn transferable mechanisms. Specifically, in the competition phase, COMET is trained with a winner-takes-all gradient allocation, encouraging the emergence of independent mechanisms. These are then re-used in the composition phase, where COMET learns to re-compose learnt mechanisms in ways that capture the dynamics of intervened environments. In so doing, COMET explicitly reuses prior knowledge, enabling efficient and interpretable adaptation. We evaluate COMET on environments with image-based observations. In contrast to competitive baselines, we demonstrate that COMET captures recognisable mechanisms without supervision. Moreover, we show that COMET is able to adapt to new environments with varying numbers of objects with improved sample efficiency compared to more conventional finetuning approaches.
\end{abstract}

\section{Introduction}
\begin{figure*}
    \centering
    \includegraphics[width=0.9\textwidth]{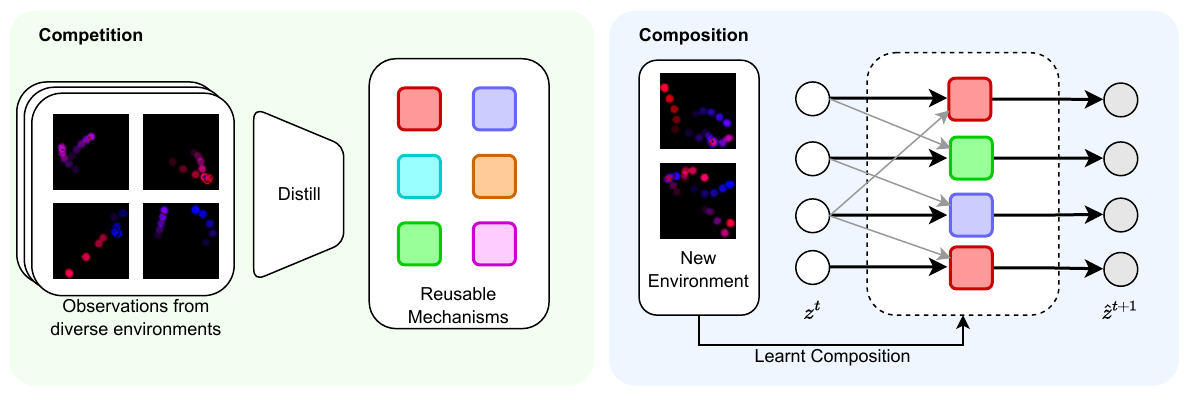}
    \caption{Illustration of the \textit{competition} and \textit{composition} training phases. In the first phase, the model learns a set of reusable mechanisms that captures interaction primitives. In the second phase, the model learns to apply these mechanisms in a new environment.}
    \label{fig:summary}
\end{figure*}

To reason about environments as rich and complex as our physical world requires the ability to learn efficiently and to flexibly adapt prior knowledge to unseen settings. Whilst humans seem able to generalise knowledge across myriad tasks and situations effortlessly, building artificial agents that can do so with minimal training data remains a significant challenge. At the heart of this challenge are crucial distinctions between what and how humans and machines learn. It has been conjectured that humans represent knowledge internally in a structured and modular way, i.e., by distilling past experience into general principles (or core knowledge)  about the world, which can be applied or selectively updated in novel settings \cite{spelke2007core, lake_ullman_tenenbaum_gershman_2017, scholkopfCausalRepresentationLearning2021,Inductive}. By contrast, current learning-based world models are mostly based on monolithic architectures, and the resulting entangled representations of the world limit the selective re-using of prior knowledge in new environments. Therefore, learning methods that afford modularity are key to world models that can adapt efficiently in diverse settings. In this paper, we address this challenge by developing a model capable of discovering a toolbox of recognisable, generalisable concepts that can be reused across different contexts.\\
\\
Recent works on object-centric world models \cite{Kipf2020Contrastive, linImprovingGenerativeImagination2020} serve as an initial step towards a structured and compositional understanding of the world. 
By decomposing the observed scene into discrete \textit{object slots}, these methods model the interaction between entities in the scene and achieve state-of-the-art results.
We argue that, just as the state representation of the scene can be factorised into object slots, the dynamics of the environment, too, can be factorised into discrete and independent \textit{mechanisms}. 
Our work is motivated by the observation that, while the overall dynamics can change across environments, we can often explain the behaviour of objects by a small set of \textit{interaction primitives} such as "A rests on top of B" and "C collides with D". 
Our understanding of novel environments is mediated by these primitives: given some observations in a previously unseen setting, one can quickly recognise the relationship between objects in terms of these interaction primitives.
As such, we posit that the ability to structurally represent different modes of interactions is crucial to flexible world models.\\
\\
Acquiring such a set of versatile mechanisms from observations without supervision presents challenges both in terms of model architecture and learning algorithm. 
While current learning methods excel at learning to predict dynamics from i.i.d. environments, learning shared modules that compositionally capture diverse environments cannot be achieved by directly applying gradient updates to the entire model. 
To this end, we argue that the ability to \textit{selectively} update the model during learning, i.e., to recognise parts of the model that are relevant to the observed data and perform modular updates, is instrumental to the emergence of discrete independent mechanisms.
In this work, we instantiate this capability by utilising a competition of experts training scheme \cite{parascandolo2018learning}, where, during training, we keep a set of independently parameterised mechanisms and only update the module which best explains the observed data.
This serves as a natural inductive bias which encourages modules to specialise in specific interaction primitives, and is consequently conducive to the emergence of reusable mechanisms.\\
\\
We introduce \textit{COmpetitive Mechanisms for Efficient Transfer} (COMET), a modular world model with the capability to explicitly disentangle modes of interactions between objects.
In contrast with prior art, COMET is able to learn discrete, abstract concepts from diverse observations, and re-use such concepts to predict the evolution of unseen environments.
We conjecture that COMET’s ability to perceive the world through the lens of structured high-level abstract concepts serves as an important step towards efficient generalisation and transfer across different task settings.
Concretely, COMET achieves this via a two-phase training procedure: i)  \textit{competition}, in which COMET learns a set of independently parameterised modules which encode interaction primitives from diverse environments; and ii) \textit{composition}, wherein COMET is trained to apply these learnt interaction primitives in novel environments (Fig. \ref{fig:summary}).\\
\\
In our experiments, we evaluate COMET on novel datasets designed to test whether models can systematically learn from multiple environment and transfer to environments with unseen dynamics. We demonstrate both quantitatively and qualitatively that COMET successfully learns reusable mechanisms, i.e. interaction primitives, from image-based observations, and is able to generalise to unseen environments by composing these mechanisms. Moreover, compared to more conventional finetuning baselines, COMET is able to explicitly make use of past knowledge and achieves improved sample efficiency during adaptation. 
\section{Related Work}
Learning internal models of the world enables decision-making agents to plan, predict and reason about the world \citep{RSSM, ha2018worldmodels, Kipf2020Contrastive}. 
As such, latent world models have attracted significant interest in recent years. These methods \citep[e.g.][]{Kipf2020Contrastive, villegas2019high, oord2018representation} in general involve learning latent representations of the state and forward prediction models. 
Our work situates in this broad context of world model learning and we focus our contribution on learning dynamics models which are factorised into composable mechanisms.
We take recent works form object-centric state representations as a starting point.
\paragraph{Object-Centric Representations}
There has been a growing interest in models that reflect the compositional nature of real-world scenarios and aim to use object-centric representations to leverage recurring features in scenes. 
Prior works have investigated unsupervised object-centric representation learning from static images \citep{burgessMONetUnsupervisedScene2019,greffMultiObjectRepresentationLearning2019,locatelloObjectCentricLearningSlot2020,linSPACEUnsupervisedObjectOriented2020,engelckeGENESISV2InferringUnordered2021}. 
Motivated by the assumption that dynamics tend to manifest themselves at the object-level \citep{hayes1979naive,baraff2001physically}, subsequent works extend this capability to video data via factorised dynamics models which operate on object-centric latent spaces. While most of these object-centric world models (OCWMs) are geared towards using temporal inputs to generate future video rollouts \citep{kosiorekSequentialAttendInfer2018,jiangSCALORGenerativeWorld2020,emamiSymmetricObjectCentricWorld2020,Kipf2020Contrastive,linImprovingGenerativeImagination2020,minGATSBIGenerativeAgentCentric2021}, some more explicitly consider their use in model-based reinforcement learning and planning \citep{wattersCOBRADataEfficientModelBased2019,veerapaneniEntityAbstractionVisual2020,singhStructuredWorldBelief2021,wuAPEXUnsupervisedObjectCentric2021}. 
In particular, graph neural networks (GNNs) are often used as a natural way to predict future states of objects and enable the modelling of interactions between objects via message passing \citep{vansteenkisteRelationalNeuralExpectation2018,tacchettiRelationalForwardModels2018,Kipf2020Contrastive,veerapaneniEntityAbstractionVisual2020,linImprovingGenerativeImagination2020,singhStructuredWorldBelief2021,sancaktarCuriousExplorationStructured2022}.
We build on these approaches by further factorising the dynamics into reusable interaction primitives.
\paragraph{Mechanism-based Models}
Our work is motivated by the conjecture that the organisation of knowledge into high-level abstract concepts is crucial to systematic generalisation \citep{Inductive}. 
This idea is similar in spirit to the \textit{Independent Causal Mechanisms} principle \citep{anticausal} and the \textit{Sparse Mechanism Shift} hypothesis \citep{scholkopfCausalRepresentationLearning2021} in the causality literature, which respectively posit that data-generating causal mechanisms operate independently from one another, and that changes in the environment can be attributed to sparse changes to such mechanisms. 
Several works \citep{VCD, huang2022adarl, lippe2022citris} have leveraged causal discovery techniques, e.g., sparsity regularisation, to learn dynamics models that are factorised into structural causal models.
\\
Similar to our approach is a class of models which represents the learned dynamics in OCWMs not as a monolithic module, but rather as a collection of independently acting mechanisms – each focusing on a different aspect of the environment’s dynamics.
\citet{becker2019switching} use a variational approach to learn to pick different transition models conditioned on the state, but is limited to linear transitions.
RIMs \citep{goyalRecurrentIndependentMechanisms2021} constitute an approach where parts of the state space are represented by independent and sparsely interacting recurrent units.
Building on this, \cite{goyalObjectFilesSchemata2020} use a GNN to model environment dynamics but reflect the concept of independent mechanisms by using different sets of GNN parameters depending on an object’s current state.  
Another approach that follows this line of work is VIM \citep{assouel22a} which considers the disentanglement of mechanisms and objects in the setting where object move independently to each other.
Closer to our method are Neural Production Systems (NPS) \citep{goyal2021neural}, another descendant of RIMs, learning a set of independent mechanisms  capturing the interaction between objects.
Our method differs from NPS in the application of competition training which, as we demonstrate empirically in Sec. \ref{sec:experiments}, is instrumental to the emergence of composable mechanisms. Furthermore, we propose a novel method for adapting to changes in the environment. 
\paragraph{Competition of Experts}
The backbone of our learning algorithm draws from mixture of experts methods \citep{jordan1994hierarchical, jacobs1990competitive, shazeer2017} and in particular from the algorithm of \citet{parascandolo2018learning}. In the context of learning independent causal mechanisms, \citet{parascandolo2018learning} demonstrate that the competition of experts algorithm induces the emergence of mechanisms that explain transformations in the data. The idea of utilising a competitive training scheme on modular model architectures has been applied on diverse settings such as lifelong learning \citep{aljundi2017expert, ostapenko2021continual}, generative models \citep{locatello2018competitive} and object-centric scene composition \citep{von2020towards}. 
Taking inspiration from this line of work, COMET uses a similar competitive training scheme as an inductive bias for disentangling modes of interaction in the setting of world model learning.
\section{COMET: COmpetitive Mechanisms for Efficient Transfer}
In this section, we present the training procedure and the architecture of COMET. The main idea of the method is to learn a set of generalisable and composable modules, which encode the different modes of interaction between objects. 
The intuition behind the approach is that, while dynamics can vary across environments, the ways in which objects or entities interact with each other can often be explained by a small number of independent rules. 
For example, in a road traffic setting, whilst the behaviour of cars can differ across different locations, the act of stopping at a red light, i.e. the interaction between cars and traffic lights, can be used to explain road behaviours in a wide range of environments. 
On a conceptual level, this can be considered as a manifestation of the ICM principle \cite{scholkopfCausalRepresentationLearning2021}. 
By having a model of \textit{how} objects could interact with each other, the task of adapting to a novel environment reduces to the learning of \textit{when} each rule should be applied. \\
\\
The goal of COMET is to discover potential interactions between objects through observed data from a diverse set of environments, and to adapt to novel environments by learning when to utilise learnt mechanisms. 
This is reflected in the training strategy of COMET, which is split into two phases: \textit{competition} and \textit{composition}. 
In the \textit{competition} phase, COMET learns a set of independent mechanisms from observed sequences using a competition of experts training scheme. 
In the \textit{composition} phase, COMET adapts to novel environments by learning a classifier that activates the correct mechanism with the correct object pair based on the state.
Our main hypothesis is that learning to apply pre-trained, reusable mechanisms in a novel environment facilitates explicit transfer of past knowledge and is hence more efficient than finetuning on new data.

\subsection{Problem Setup and Model Architecture}
COMET learns from a dataset of observed sequences $\{\mathbf{x}^{1:T}\}_N$.
In this work we focus on learning from observations without actions, which is commensurate with the settings in similar works \cite{linImprovingGenerativeImagination2020, jiangSCALORGenerativeWorld2020}, although the framework presented here can be readily extended to action-conditioned world models.
Importantly, these sequences are sampled from environments with varying dynamics where objects can exhibit different behaviours. In order to model interactions between objects, we assume that each observation, $\mathbf{x}^t$, can be factorised into latent object-slots, $\{\mathbf{z}^t_0, \mathbf{z}^t_1, ..., \mathbf{z}^t_K\}$, where the subscript denotes the object-id. These object representations can be based on ground-truth state information or obtained from object-centric encoders \cite{burgessMONetUnsupervisedScene2019,locatelloObjectCentricLearningSlot2020,engelckeGENESISV2InferringUnordered2021}.\\
\\
COMET consists of two main components, the \textit{mechanisms} and the \textit{composition module}. \emph{Mechanisms} contains $M$ independently parameterised feedforward networks, $f_{mech}^m: \mathbb{R}^{2d} \rightarrow \mathbb{R}^d$, with parameters $\theta_m$, where $d$ is the dimension of the object representations.
Each mechanism predicts updates to all objects at every timestep, given the state of the object itself and another context object:
\begin{equation}
    \Delta\mathbf{z}_i^t(m,j) = f_{mech}^m([\mathbf{z}^t_i \oplus \mathbf{z}^t_j]),
\end{equation}
where $\oplus$ denotes concatenation, $i$ is the index of the object to be predicted and $j$ is the index of the \textit{context} object, i.e. the object with which object $i$ interacts.
The mechanisms are trained during the \textit{competition} phase where each mechanism learns to specialise to cover a particular mode of interaction between objects.\\
\\
In order to predict transitions for objects using the trained mechanisms, the composition module picks the relevant context object and the active mechanism.
The composition module predicts the mechanism-context pair based on the observed state:
\begin{equation}
    (\hat{m}_i, \hat{j}_i) = f_{comp}(\mathbf{z}^t_i, \mathbf{z}^t_{1:K}),
\end{equation}
where $(\hat{m}_i, \hat{j}_i)$ is the predicted mechanism-context pair for object $i$ and $\mathbf{z}_{1:K}$ is the representation of all objects. 
$f_{comp}$ can be considered as a classifier with $M \times K$ classes, where $M$ is the number of mechanisms and $K$ is the number of objects in the scene.
In the \textit{composition} phase, COMET adapts to a new environment given a small number of observed sequences $\{\tilde{\mathbf{x}}^{1:T}\}_{\tilde{N}}$ by updating the composition module. 

\subsection{Phase 1: Learning Reusable Mechanisms via Competition}
\label{sec:competition}
\begin{figure*}
    \centering
    \includegraphics[width=0.9\textwidth]{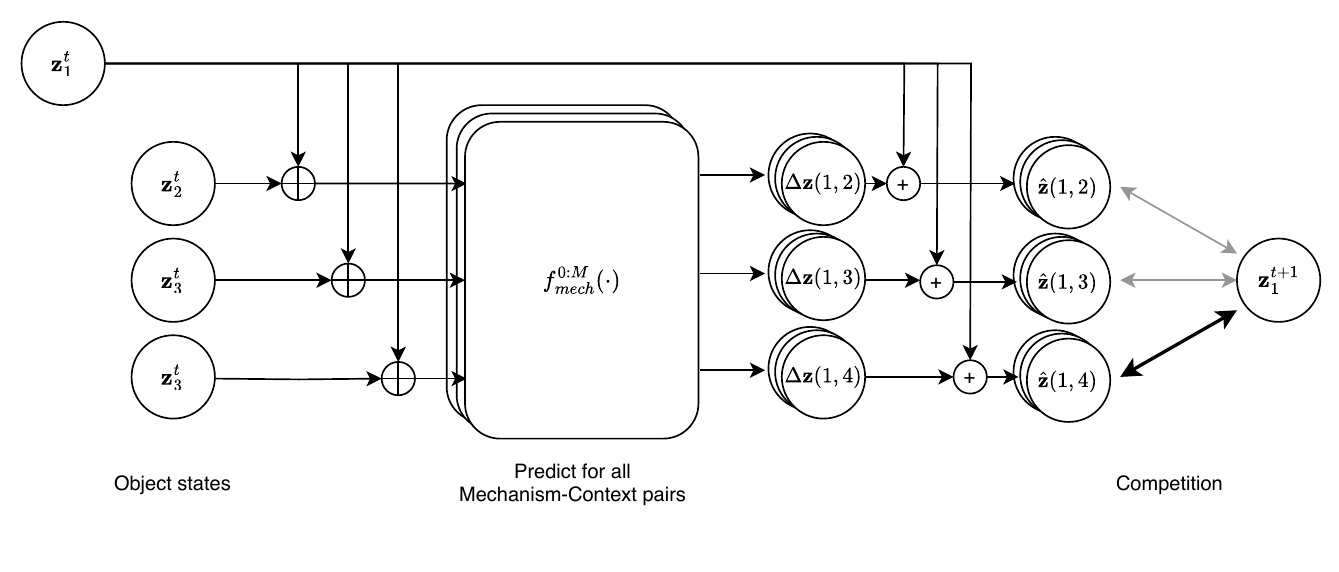}
    \caption{In the \textit{competition} phase, predictions are made using all possible mechanism-context pairs for each object. Gradients are only allocated to the mechanism-context pair which produces the most accurate prediction. This encourages specialisation within the mechanisms and enables learning from environments with varying dynamics. The figure describes the prediction step for a single object.}
    \label{fig:competition}
\end{figure*}  
The aim of the \textit{competition} phase is to train the mechanisms in a way such that each mechanism specialises in a particular mode of interaction between objects. Since the training dataset is sampled from multiple environments with varying dynamics, the training procedure needs to recognise shared mechanisms regardless of context. Taking inspiration from \citet{parascandolo2018learning}, we apply a competitive training scheme in the setting of dynamics learning. Concretely, for each object, the model makes predictions using all possible mechanism-context pairs in parallel. Comparing the predictions, we update the mechanism-context pair with the most accurate prediction for a given object. This is illustrated in Fig.~\ref{fig:competition}. Given a state transition pair, $(\mathbf{z}^t_{1:K}, \mathbf{z}^{t+1}_{1:K})$, the loss function can be written as:
\begin{equation}
    \label{competition_loss}
    \mathcal{L}(\theta_{1:M}) = \sum_{i=0}^K \min_{m,j} \left[d\left(\mathbf{z}_i^t + \Delta\mathbf{z}^t_i(m,j), \mathbf{z}_i^{t+1}\right)\right],
\end{equation}
where $d$ is a function that measures the prediction error. There are several sensible choices for the distance function $d$. 
In this work, we find that using Euclidean distance is sufficient. 
Other potential choices include a contrastive loss \cite{Kipf2020Contrastive} or ELBO loss \cite{RSSM} for training representation and dynamics model jointly, and sparsity-regularised loss \cite{VCD, huang2022adarl} for causal structure discovery. 
Importantly, when performing back-propagation on this loss function, only the parameters of the competition winner are updated. 
This independent update of parameters encourages \textit{specialisation} among the mechanisms and facilitates environment-agnostic learning. 
Intuitively, insofar as a particular mode of interaction is concerned, a specialised mechanism will outperform other mechanisms and hence win more data as training progresses. This triggers a positive feedback loop which allows the specialised mechanism to further improve its accuracy.
Moreover, the competition procedure ensures the selection of the correct context objects during training since accurate predictions are only possible with the relevant context information.\\
\\
In practice, one of the main failure modes in this training phase is a case where a single mechanism wins the competition most of the time, which hinders the proper disentanglement of dynamics.
Empirically, we find that the robustness of the training dynamics can be significantly improved by 1) adding a warm-starting phase during which gradients are equally distributed; and by 2) increasing the time horizon for mechanism selection which improves the robustness of the training dynamics.
This means that a mechanism-context pair is picked only if it has the best performance over multiple consecutive time steps. 
In App.~\ref{app:ablation_studies}, we discuss this further with ablation studies.

\subsection{Phase 2: Learning to Compose Mechanisms in New Environments}
With a trained set of mechanisms, COMET adapts to new environments by training the composition module on new data.\footnote{Here, we focus on directly reusing mechanisms new environments. We leave the exploration of systems that can instantiate new mechanisms to future work. See discussion in Sec. \ref{sec:limitations}.} The composition module acts as a classifier which picks the optimal mechanism-context pair for each object in the scene. Given the state of object $i$, for each mechanism-context pair, we compute the confidence score:
\begin{equation}
    c^t_i(m, j) = f_{conf}^m([\mathbf{z}^t_i \oplus \mathbf{z}^t_j]),
\end{equation}
where $f_{conf}^m$ is an independently parameterised MLP for mechanism $m$, i.e. each mechanism has a corresponding $f_{conf}^m$. We predict a categorical distribution over all mechanism-context pairs by taking the softmax over the confidence scores for object $i$ at time step $t$.\\
\\
Given a small number of observation sequences in a new environment, we obtain the best performing mechanism-context pair $(m^*, j^*)_i^t$ for each object at each time step by investigating which pair minimises the loss function of the competition scheme in Eq. \ref{competition_loss}. 
These then serve as the target labels for the classifier. 
The composition module is then trained using the negative log-likelihood loss. 
At test time, together with the trained mechanisms, COMET predicts the next state of each object by first picking the mechanism-context pair with the highest score and then feeding the chosen context object through the chosen mechanism.
In the following section, we show empirically that this two-step training procedure is able to learn meaningful mechanisms from image observations and can compose mechanisms in novel environments efficiently.
\section{Experiments}
\label{sec:experiments}
In this section, we demonstrate that COMET is able to disentangle different modes of interaction between objects and can efficiently reuse learnt mechanisms during adaptation. We evaluate, both quantitatively and qualitatively, the performance of COMET on image-based environments. Concretely, the experiments focus on whether COMET can learn reusable and recognisable mechanisms (Sec. \ref{sec:disentanglement_of_mechanisms}) and whether re-composing learnt mechanisms is more sample-efficient compared to finetuning approaches (Sec. \ref{sec:adaptation_efficiency}).
\subsection{Experimental Setup}
\paragraph{Baselines.}
We evaluate COMET against two competitive baselines, C-SWM \cite{Kipf2020Contrastive} and Neural Production Systems (NPS) \cite{goyal2021neural}. 
C-SWM learns a world model from observation via contrastive learning with a GNN-based dynamics model. 
Similar to COMET, C-SWM operates on an object-factorised representation and achieves state-of-the-art prediction accuracy. 
COMET further disentangles the interactions between objects as independent mechanisms rather than learning a monolithic model that captures all interactions. 
In Sec. \ref{sec:adaptation_efficiency}, we show that this disentanglement is conducive to sample-efficient adaptation and facilitates interpretable transfer.\\
\\
We also compare against NPS, which learns to capture object interactions as independent mechanisms. Architecturally, NPS is similar to COMET, except that the mechanism-context pair is selected using dot-product attention \cite{vaswani2017attention} which is trained jointly with the mechanisms. In contrast, COMET deploys a competitive training scheme which allows the model to recognise shared mechanisms across environments.
Moreover, in Sec. \ref{sec:disentanglement_of_mechanisms}, we show that competition serves as a strong inductive bias that enables the emergence of generalisable mechanisms. We provide more model details and further discussions around the baselines in App. \ref{app:training_details_and_baselines}.
\begin{figure*}[t]
    \centering
    \includegraphics[width=\textwidth]{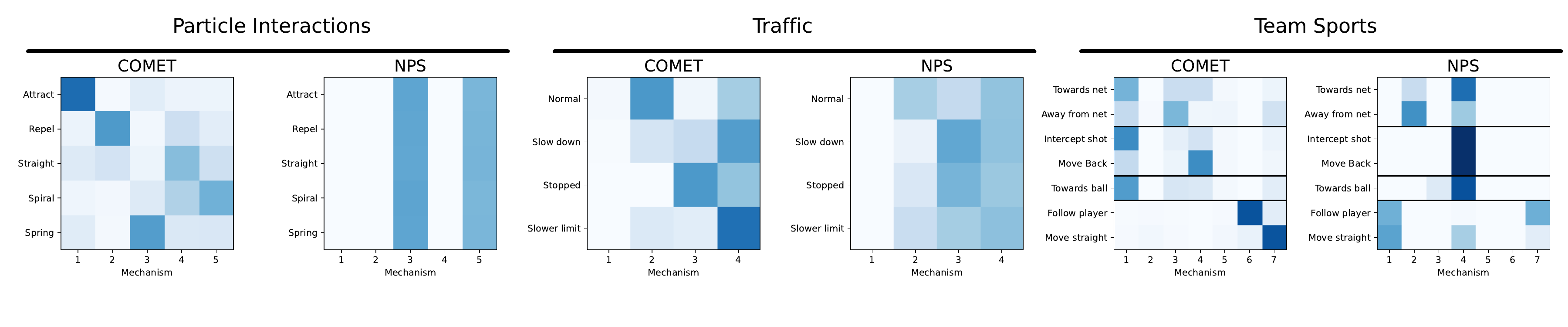}
    \caption{
    Disentanglement plots showing the correlation between mechanisms chosen by the models and ground-truth interaction modes. In the ideal case, the matrices should look like permutation matrices. Here, COMET is able to learn disentangled mechanisms that correspond to ground-truth behaviours in all three domains, as indicated by the fact that each interaction mode has one main corresponding learnt mechanism. In contrast, NPS does not exhibit the same structure.
    }
    \label{fig:disentanglement}
\end{figure*}
\begin{figure*}[t]
    \centering
    \includegraphics[width=\textwidth]{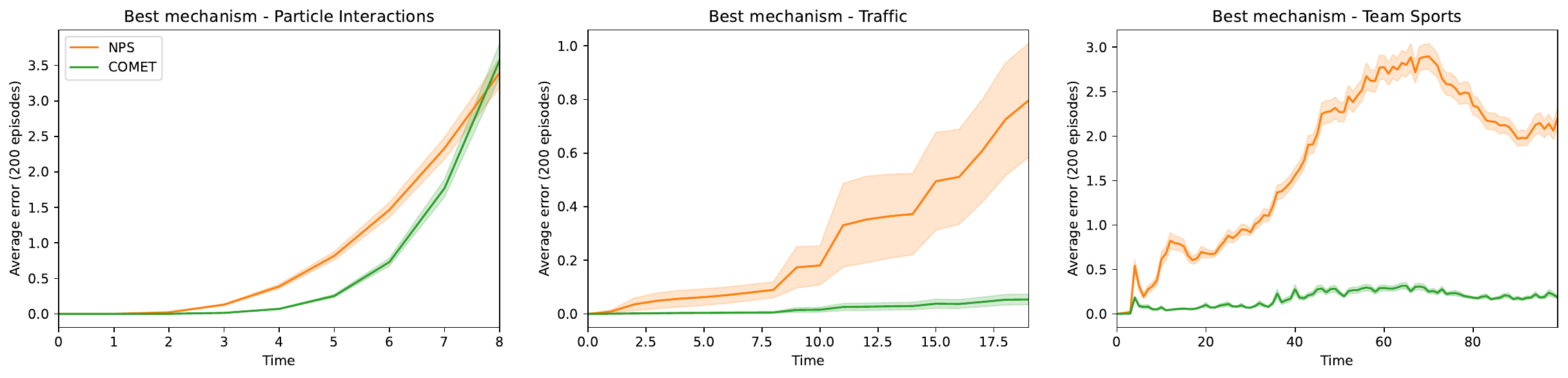}
    \caption{Rollout errors (lower is better) in unseen environments with optimal mechanism selection. Shaded areas indicates the standard error of the mean. The lower errors indicate that COMET mechanisms can be readily reused across environments \textit{without} finetuning.}
    \label{fig:best_mechanism}
\end{figure*}
\paragraph{Datasets.}
We evaluate COMET on three problem domains: \textit{Particle Interactions}, \textit{Traffic}, and \textit{Team Sports}. For each of these domains, we define a set of environments where objects can exhibit different behaviours. These environments are designed to test whether COMET can extract meaningful mechanisms and adapt to unseen environments via composition. The \textit{Particle Interactions} dataset consists of coloured particles that can interact with each other in different ways such as attraction and repulsion. Environments are defined by a combination of rules such as "red particles repel each other". The \textit{Traffic} dataset contains observation sequences of traffic scenarios generated with the CARLA simulator \cite{Dosovitskiy17}. Here, the environments are defined by traffic rules that apply to different vehicles such as "blue cars do not need to stop at red lights". The \textit{Team Sport} domain consists of a simulated generic hockey game where players can perform different actions such as moving towards the puck or dribbling the puck towards the opponent goal, generated by the STS2 simulator \cite{sts2_ea_2020}. In this setting, differences in environment amounts to the different behaviours of the players, e.g. players in some environments might tend to take more aggressive actions. This is a particularly challenging dataset for COMET as it violates the assumption that all interactions are binary. Observations are masked RGB images for each object, except for the \textit{Team Sports} domain, where state-based observations are more naturally suited. Details of the datasets are provided in App. \ref{app:experiment_details}.

\subsection{Disentanglement of Mechanisms}
\label{sec:disentanglement_of_mechanisms}
\begin{figure*}[h]
    \centering
    \includegraphics[width=0.9\textwidth]{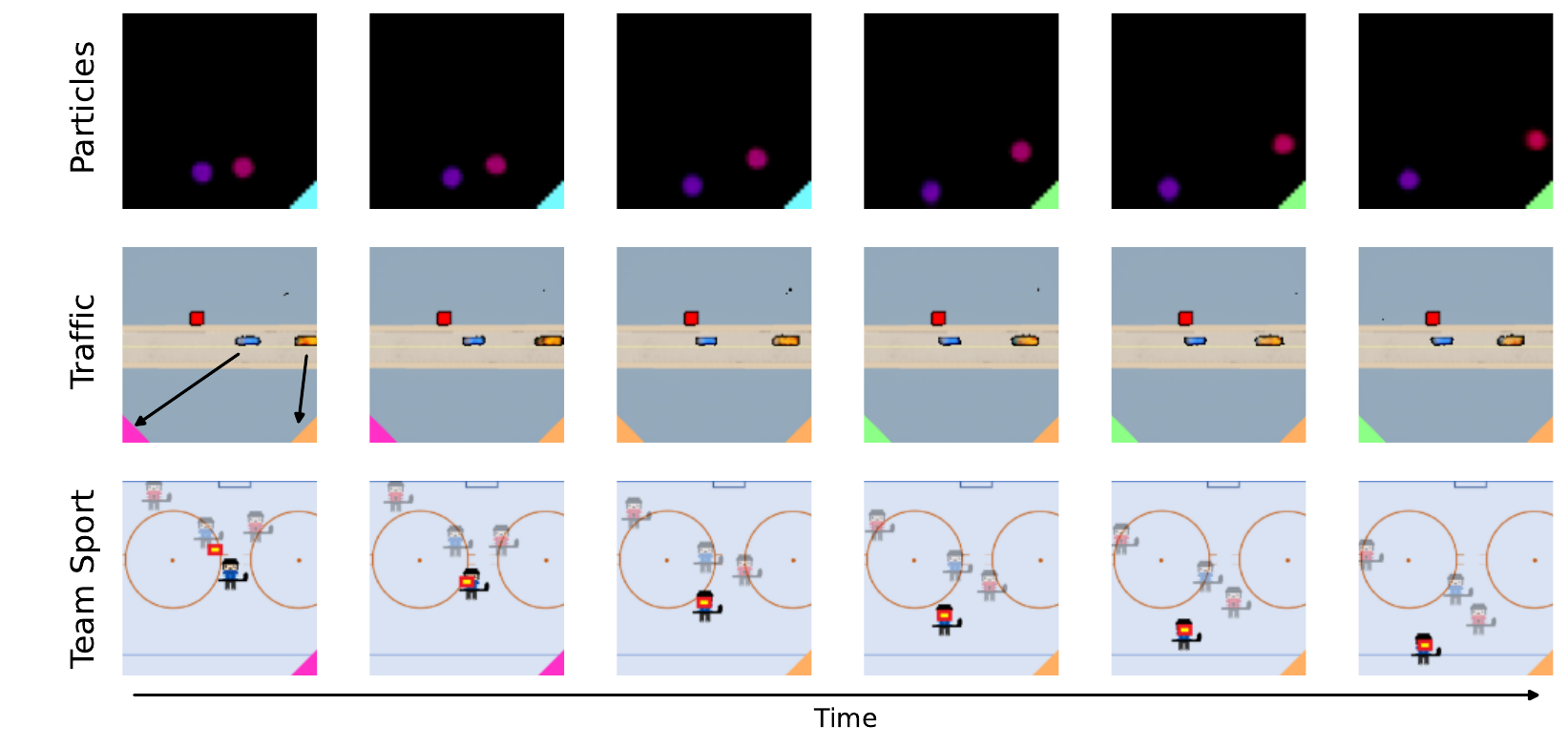}
    \caption{Qualitative rollouts. The colour of the tabs on the bottom of each frame indicates the 'winning' mechanism at each time step. Across all environments, the competition winner changes as the underlying interaction mode changes. \textbf{Top}: The particles repel each other when they are close (blue) and moves independently when they are apart (green). \textbf{Middle}: In this traffic environment, the orange car obeys a slower speed limit and always pick the slow mechanism (orange). The blue car approaches the red light with normal driving (pink) $\rightarrow$ slow down (orange) $\rightarrow$ stop (green). Note that the orange mechanism is used as slow driving for both cars. \textbf{Bottom}: The player first wait to receive the ball (pink) and the moves towards opponent goal when in pocession of the ball (orange).}
    \label{fig:rollout_example}
\end{figure*}

We investigate the emergence of recognisable mechanisms from competition. Here, both COMET and NPS are trained on a mixture of environments in each domain. We obtain the ground-truth labels for the object interactions and use these to directly investigate whether the learnt mechanisms correspond to actual interactions without any supervision. These labels are not accessible to the models during training. Fig. \ref{fig:disentanglement} shows the correlation between the active ground-truth interactions and the winning mechanisms in the competition process in the different domains. COMET achieves successful disentanglement and learns mechanisms that corresponds to the ground-truth interactions. In the \textit{particle interaction} dataset, COMET recovers the ground-truth mode of interactions between particles. Similarly, in the \textit{Traffic} and \textit{Team Sports} datasets, COMET learns mechanisms that model interaction primitives such as stopping before a red light and running towards the opponent goal. \footnote{In the team sports dataset, we observe that mechanism 1 is 'overloaded' in the sense that it represents different behaviours for different objects, i.e. for the attacking team, it models 'move towards net' and for the defending team, it models 'intercept shot'. Crucially, for each object type (indicated by the horizontal divider lines in Fig. \ref{fig:disentanglement}), different behaviours are captured with different mechanisms. This is not the case for NPS.} 
Fig. \ref{fig:rollout_example} qualitatively illustrate that the 'winning' mechanisms switches as the underlying interaction type changes.
In contrast, the mechanisms learnt by NPS show no correspondence with the ground-truth interactions. 
This is likely because NPS cannot learn from a mixture of environments with varying dynamics as it employs a simple dot-product attention for picking mechanisms during training. To this end, COMET's ability to learn from diverse environments is uniquely afforded by the competition scheme which assigns relevant data to update each mechanism.\\ 
\\
Central to our hypothesis is the notion that good mechanisms are \textit{reusable} across environments.
To this end, we perform rollout in \textit{unseen} environments by picking the optimal mechanism-object pair, $(m^*, j^*)$, effectively bypassing the confidence module. This offers insights to whether the learnt mechanisms can generalise to new settings.
We perform this to the mechanisms learnt with COMET and NPS. Fig. \ref{fig:best_mechanism} shows the rollout errors with optimal mechanism selection.
COMET mechanisms performs significantly better than NPS mechanisms, meaning that, when picked correctly, COMET mechanisms can be directly used in new environment without any finetuning. This corroborates our claim that competition training is conducive to the emergence of generalisable and reusable mechanisms. 
\subsection{Adaptation Efficiency}
\label{sec:adaptation_efficiency}
\begin{figure*}[h]
    \centering
    \includegraphics[width=\textwidth]{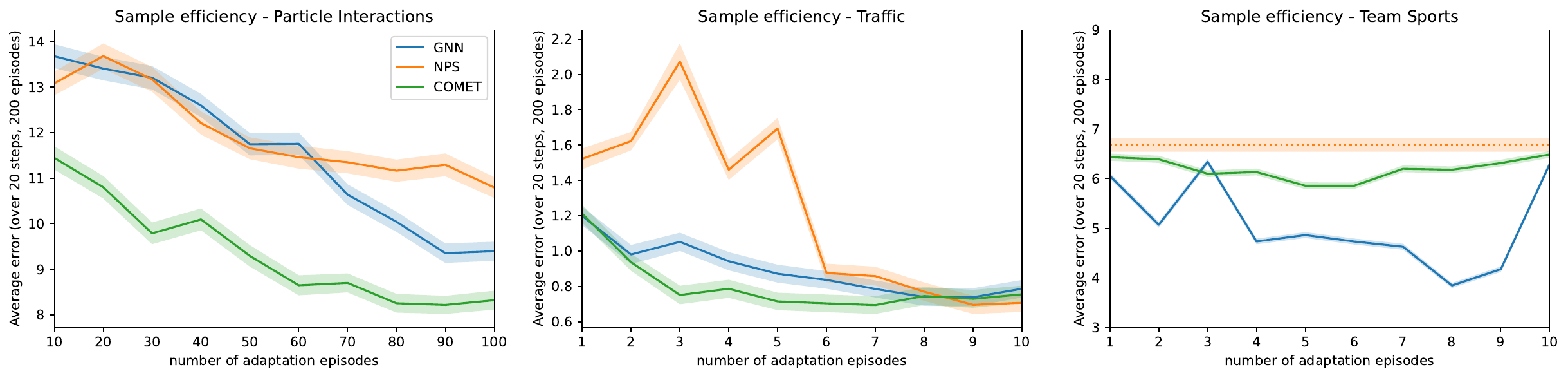}
    \caption{The average rollout error in an unseen environment with different amount of observed data in the new environment (lower is better). In all environments, all models eventually converge to similar errors given enough data. We show this explicitly in App \ref{app:extra_rollouts} and offer further discussion. In terms of sample efficiency, in the \textit{Particle Interactions} and \textit{Traffic} domains, COMET is able to achieve lower errors with few adaptation episodes. This means that COMET can learn to use the correct mechanisms with a small amount of data, thus corroborates our hypothesis that composing learnt mechanisms enables sample-efficient transfer. In the \textit{Team Sports} domain, NPS is not able to generate stable rollouts with the amounts of adaptation episodes shown in the plots. The dotted line indicates the performance of NPS when trained with a large amount of data. Shaded areas represent the standard errors of the mean.
    }
    \label{fig:adaptation_plots}
\end{figure*}
One of our main hypotheses is that learning to compose learnt mechanisms leads to data-efficient adaptation. 
For each domain, we train all of the models on a mixture of environments and adapt the models to unseen environments. COMET adapts by training the composition module, whereas the baselines adapt by finetuning the entire model on new data. Fig. \ref{fig:adaptation_plots} shows the prediction performance of the models when trained on different amounts of observations in the new environment. In general, all models improve as the amount of available data increases. In the \textit{Particle Interactions} and \textit{Traffic} domains, while all models eventually perform comparably given sufficient data, here we see that COMET outperforms the baselines in the low-data regime, illustrating that explicitly reusing learnt mechanisms results in improved sample efficiency compared to gradient-based finetuning. In the \textit{Team Sport} domain, C-SWM performs better than COMET and NPS. We hypothesise that this is because both COMET and NPS only model binary interactions between objects whereas C-SWM uses a GNN-based transition model that can take into account the entire scene. We further discuss this limitation in Sec. \ref{sec:limitations}. Nonetheless, under the same binary interactions restriction, COMET significantly outperforms NPS, which overfits to the adaptation data quickly and is not able to generate stable rollouts.

\subsection{Limitations}
\label{sec:limitations}
\paragraph{Object Encoders.}
In our experiments, we focus our analysis on the dynamics models and opt for simple CNN encoders with ground-truth object segmentation masks as the representation backbone. The application of the model on more visually complex environments without segmentation masks would require the use of object-centric encoders or other segmentation techniques \cite{elsayed2022savipp, Genesisv2, locatelloObjectCentricLearningSlot2020}.
\paragraph{Binary Interactions.}
The model architecture for the mechanisms implies that the model can only capture \textit{binary} interactions. While this is a good approximation for many interacting systems, this hinders the model's ability to capture more complex dynamics. As seen in the Team Sports experiments, while COMET can disentangle meaningful mechanisms, picking correct mechanisms may require information from the entire scene, leading to suboptimal prediction accuracy. Considering $n$-ary interactions naively would require comparing exponentially many options in the competition phase. We leave the exploration of higher-order interactions to future work.
\paragraph{Adapting Mechanisms.}
As it stands, COMET lacks the ability to update the mechanisms during \textit{composition}, which limits the model's ability to adapt to environments with completely new interactions. Designing mechanism-based models that can instantiate new mechanisms as it encounters new data offers an exciting avenue of research, with important implications on life-long learning, where an agent can improve its understanding of the world through a growing set of reusable mechanisms. This is beyond the scope of the present investigation. Nonetheless, we believe the method presented here will be instrumental in the development of such systems.
\section{Conclusion}
In this paper, we introduce COMET, a structured world model which encodes discrete abstract mechanisms explicitly from observations. 
This is achieved through training the model on a set of environments with diverse dynamics using a winner-takes-all competition scheme. 
This enables the model to perform \textit{selective} updates during the training phase, a central capability which facilitates the emergence of recognisable and reusable mechanisms.
We show experimentally that the proposed method is indeed able to disentangle shared mechanisms across different environments from image observations, and thus enables sample-efficient and interpretable adaptation to novel situations.
While the presented results are competitive on the tasks considered, we see our main contribution as a conceptual one, serving as a step towards world models exhibiting a structured understanding of the world.
Looking forward, we believe that the method developed here opens up several promising avenues of research, such as designing agents that learn a growing repertoire of re-usable interaction behaviours and agents that explore the world through the lens of mechanism discovery.

\section*{Acknowledgements}
This research was supported by an EPSRC Programme Grant (EP/V000748/1). The authors would like to acknowledge the use of the SCAN facility
\medskip

\small

\bibliography{ref}
\bibliographystyle{plainnat}

\newpage
\appendix
\newpage
\appendix
\section{Model Details and Baselines}
\label{app:training_details_and_baselines}

\subsection{Representation Model}
\label{app:representation_model}
In our experiments, all models are trained with image input in the \textit{Particle Interactions} and \textit{Traffic} domains. At each timestep, the observation is given as a set of image segmentations - one segmentation frame for each object in the scene. These are then transformed into object embeddings using a pre-trained CNN encoder. Each segmented frames are encoded separately, giving latent representations of each object. To allow for the capture of dynamic visual information, we concatenate two segmentation frames from consecutive time steps along the channel dimension. 
We pre-train the representation model using the C-SWM \cite{Kipf2020Contrastive} framework, which uses uses a contrastive loss signal in latent space. The hyperparameters used during training of the encoder are equivalent to those reported in \cite{Kipf2020Contrastive}.\\
\\
For the \textit{Team Sports} domain, the ground-truth states of the players and the ball is used directly.
\subsection{COMET Architecture}
The architecture of COMET consists of two parts, the \textit{mechanisms} and the \textit{composition module}, as described in the main text.
The first part, the set of mechanisms $f_{mech}^{0:M}(\cdot)$, operates on the concatenated representations of two objects, the primary object and the context object.
The mechanisms are implemented as feed-forward networks. 
The architecture of a single mechanism is shown in Table \ref{tab:mechanism_architecture}.
The composition module itself consists of a set of independent feed-forward confidence networks, one for each mechanism. Each such independent network receives the concatenated representations of two objects as input, similar to the mechanisms, and outputs a scalar confidence score. The architecture of a single confidence network is shown in Table \ref{tab:confidence_network_architecture}. The source code for COMET will be released with a camera ready version of the paper.

\begin{table}[]
\centering
\scalebox{0.8}{
\begin{tabular}[t]{@{}llll@{}}
\toprule
 & Layer Type & in Features & out Features \\ \midrule
\multicolumn{1}{l|}{1} & \multicolumn{1}{l|}{Linear} & 2 * 8 & 300 \\
\multicolumn{1}{l|}{2} & \multicolumn{1}{l|}{ELU} &  &  \\
\multicolumn{1}{l|}{3} & \multicolumn{1}{l|}{Linear} & 300 & 300 \\
\multicolumn{1}{l|}{4} & \multicolumn{1}{l|}{ELU} &  &  \\
\multicolumn{1}{l|}{5} & \multicolumn{1}{l|}{Linear} & 300 & 8 \\ \bottomrule
\end{tabular}
}
\vspace{3mm}
\caption{Architecture of a single mechanism. The mechanism operates on the concatenation of two object embeddings and predicts state updates for the first input object.}
\label{tab:mechanism_architecture}
\end{table}
\begin{table}[]
\centering
\scalebox{0.8}{
\begin{tabular}{@{}llll@{}}
\toprule
 & Layer Type & in Features & out Features \\ \midrule
\multicolumn{1}{l|}{1} & \multicolumn{1}{l|}{Linear} & 2 * 8 & 300 \\
\multicolumn{1}{l|}{2} & \multicolumn{1}{l|}{ELU} &  &  \\
\multicolumn{1}{l|}{3} & \multicolumn{1}{l|}{Linear} & 300 & 300 \\
\multicolumn{1}{l|}{4} & \multicolumn{1}{l|}{ELU} &  &  \\
\multicolumn{1}{l|}{5} & \multicolumn{1}{l|}{Linear} & 300 & 1 \\ \bottomrule
\end{tabular}
}
\vspace{3mm}
\caption{Architecture of a single confidence network. There is one confidence network per mechanism and it operates on the concatenation of two object embeddings to predict a score for the given context object and associated mechanism.}
\label{tab:confidence_network_architecture}
\end{table}
\vspace{-2mm}
\subsection{Baselines} 
\subsubsection{Contrastive Learning of Strutured World Models}
Contrastive Learning of Strutured World Models (C-SWM) \cite{Kipf2020Contrastive} is a state-of-the-art object-centric world model. At its core is a GNN transition model that takes object representations as its nodes and is assumed to be fully-connected. We use the official implementation of the model \footnote{\url{https://github.com/tkipf/c-swm}} but replace the default encoder with the encoder presented in Appendix \ref{app:representation_model}. The default encoder of C-SWM decomposes a single RGB-frame into a set of objects whereas our encoder operates on a set of object segmentation frames. Other than that, we use the original training procedure and model architecture as presented in \cite{Kipf2020Contrastive}. This means that the only difference between the C-SWM implementation and the COMET implementation in our experiments is the use of a GNN transition model.

\subsubsection{Neural Production Systems}
Neural Production Systems (NPS) \cite{goyal2021neural} is an object-factorised world model that, like COMET, operates with a set of mechanisms instead of a monolithic transition model architecture. Furthermore, NPS also considers sparse contexts, i.e. an object’s transition is conditioned on the state of at most one other object in the observation. 

The most prominent difference between NPS and our approach is that NPS jointly learns the mechanism selection, as predicted by a dot-product attention layer, and the mechanisms themselves, whereas COMET decouples the mechanism learning and the selection of mechanisms. This limits NPS’ capability of being trained on multiple environments, as different mechanisms may be active in different environments given the same state. 
In contrast, the competition of experts setup of COMET produces predictions for all mechanisms and updates the one with the lowest prediction error. Thus, the only requirement for COMET is that during the competition stage, enough mechanisms exist to capture all environment dynamics. 
The selection of mechanisms is separately trained during the environment-specific composition phase.

In our experiments, the NPS transition model replaces the GNN transition model of the C-SWM architecture, and uses the same representation model presented in App. \ref{app:representation_model}. 
Since no code is provided with the paper, we implemented our own version based on the detailed algorithms provided in the appendix of the original paper \cite{goyal2021neural}. 
We verified the implementation by confirming that model performance is on par with the results presented in the original paper when evaluated on similar environments.
\subsection{Hyperparameters}
During training for all of the models, the batch size is set to 1024 and we use the Adam \cite{kingma2014adam} optimiser with a learning rate of 1e-4. Object embeddings are set to be 8-dimensional.

\subsection{Compute Resources}
All experiments were run on a cluster containing a mixture of Nvidia Quadro RTX 6000 and Nvidia Tesla V100 accelerators. Each training run involved a single GPU only. All training runs are finished within 24 hours. 

\section{Datasets}
\label{app:experiment_details}
We evaluate COMET and the baselines on three problem domains, \textit{Particle Interactions}, \textit{Traffic} and \textit{Team Sport}. In each of the domains, we define a set of environments with varying dynamics. Below we provide the details of the domains.
\subsection{Particle Interactions}
In the \textit{Particle Interactions} domain, each environment is defined by a set of rules. Each of these rules consists of an interaction and a condition (i.e. when the interaction should happen). The particles interact in one of 5 modes of interactions: \textit{straight line}, \textit{repulsion}, \textit{attraction}, \textit{spring} and \textit{spiral towards centre}. Moreover, each particle is coloured. We pick 6 randomly generated environments as the training set, summarised in Table \ref{tab:environments}. The training set consists of 2000 sampled episodes from each of these environments, each with 50 frames.\\
\\
For the adaptation experiments, COMET adapts to an unseen environment with 4 particles, with the \textit{spring} interaction if the particles are of the same colour and \textit{repel} otherwise.
\begin{table}[h]
\centering
\scalebox{0.8}{
\begin{tabular}{@{}llll@{}}
\toprule
 & Condition & Interaction\\ \midrule
\multicolumn{1}{l|}{Environment 1} & close together & repulsion \\
\multicolumn{1}{l|}{ } & otherwise & straight line\\
\midrule
\multicolumn{1}{l|}{Environment 2} &  same colour & spring \\
\multicolumn{1}{l|}{ } & close together & attraction\\
\multicolumn{1}{l|}{ } & otherwise & straight line\\
\midrule
\multicolumn{1}{l|}{Environment 3} & same colour & repulsion \\
\multicolumn{1}{l|}{ } & opposite colour & spring\\
\multicolumn{1}{l|}{ } & otherwise & straight line\\
\midrule
\multicolumn{1}{l|}{Environment 4} & same colour  & attraction\\
\multicolumn{1}{l|}{ } & is blue & spiral towards centre\\
\multicolumn{1}{l|}{ } & is red & spiral towards centre\\
\multicolumn{1}{l|}{ } & otherwise & straight line\\
\midrule
\multicolumn{1}{l|}{Environment 5} & same colour & repulsion \\
\multicolumn{1}{l|}{ } & otherwise & spring\\
\midrule
\multicolumn{1}{l|}{Environment 6} & always & spiral towards centre\\

\bottomrule
\end{tabular}
}
\caption{Environment details for particle interactions domain.}
\label{tab:environments}
\end{table}

\subsection{Traffic}
The \textit{traffic} domain is simulated using the CARLA simulator \citep{Dosovitskiy17}. License information can be found on \url{https://github.com/carla-simulator/carla}. Here, the objects are two cars, one orange and one blue, and a traffic light. In different environments, they follow different traffic rules. We train the models on three environments, 1. cars drive at normal speeds and stop at red traffic light, 2. cars drive under a much slower speed limit and stop at red traffic light, and 3. cars ignore traffic lights. In the adaptation environment, only the orange car drives at a slow speed limit. As such, the model needs to explain new behaviours such as the blue car slowing down to avoid collision with the orange car in front. The training set has 2000 episodes from each environment. The length of each episode depends on the exact scenario, which ends if both cars are out of frame or have stopped moving over 7 timesteps.
\subsection{Team Sport}
The \textit{Team Sport} domain is simulated using a modified version of STS2 \citep{sts2_ea_2020} (\url{https://github.com/electronicarts/SimpleTeamSportsSimulator}). As the domain comes with only a single player type, we define additional player types according to heuristics:
\subsubsection{Simple Player}
If the simple player is in control of the ball it moves towards the opponent goal. If other team players are closer to the opponent goal it tries to pass the ball to them. If a simple player is in the vicinity of the opponent goal and in control of the ball, it tries to score a goal. If a different player of the own team is in control of the ball, the simple player moves towards the opponent goal. If the ball is in control of the opponent team, the simple player moves towards the midway point between the own goal and the control player to block any attempted shot. If the ball is in the air due to a pass in progress, the simple player will move towards the ball
\subsubsection{Defensive Player}
The defensive player inherits all offensive behaviour from the simple player. If the defensive player is not in control of the ball, it moves towards the own goal to diffuse any counter attacks.
\subsubsection{Shy Player}
The shy player inherits all behaviour from the defensive player but instead of moving towards the opponent goal if in control of the ball, it moves away from the opponent goal.

For all player types, we define a specific course of action as being a single mechanism. Player types may share some mechanisms (e.g. moving towards the opponent goal) but differ in others. After executing certain actions such as shooting, passing, receiving, or tackling, players are unable to perform another action within a set number of time steps. This accounts for human players also requiring a certain reaction and orientation time after executing such actions. 

We defined separate mechanisms for the ball which is modeled as a separate object. If in possession, the ball traces the control player and if in the air as a result of a pass or shot, the ball follows a straight line at constant speed until it is either caught by a player or bounces off a wall to change its direction. A list of all mechanisms is depicted in Fig. \ref{fig:disentanglement}.

We collect a total of 5,957 episodes over four different team configurations for training and evaluate our model and baselines on a configuration of two simple players in each team, a setting which was not part of the training configurations. 

The state space consists of three one-hot flags indicating home team, away team, or ball. Further, we include two one-hot variables indicating which team is currently in possession of the ball (if any), a one-hot variable indicating whether the player itself is in control of the ball, a countdown timer denoting for how many time steps the player is unable to execute an action after having executed an action shortly before, as well as position and velocity information. Dimensions that apply only to players but not to the ball (e.g. the countdown timer) are present in the ball state but set to zero.

\section{Stability of Mechanism Disentanglement}
\begin{figure}[h]
  \begin{center}
    \includegraphics[width=0.38\textwidth]{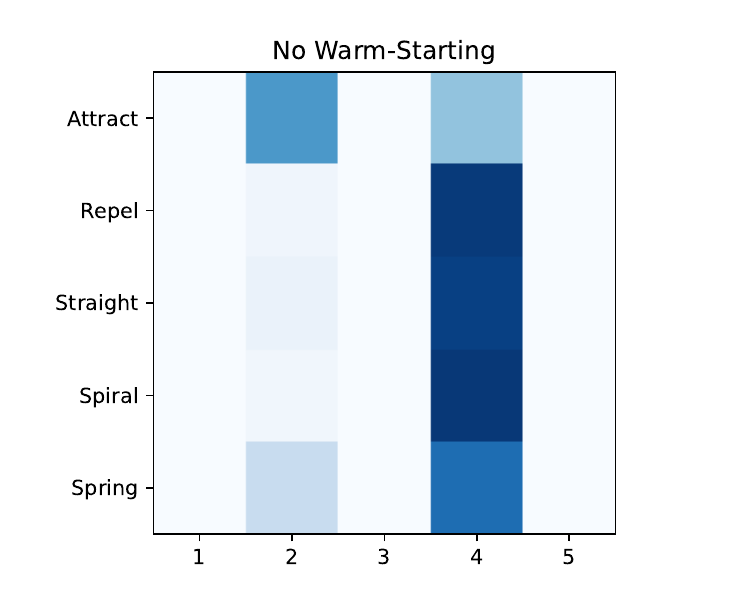}
  \end{center}
  \caption{Disentanglement plot for training without warm-starting.}
  \label{fig:no_warm_start}
\end{figure}
\begin{figure*}[h]
    \centering
    \includegraphics[width=\textwidth]{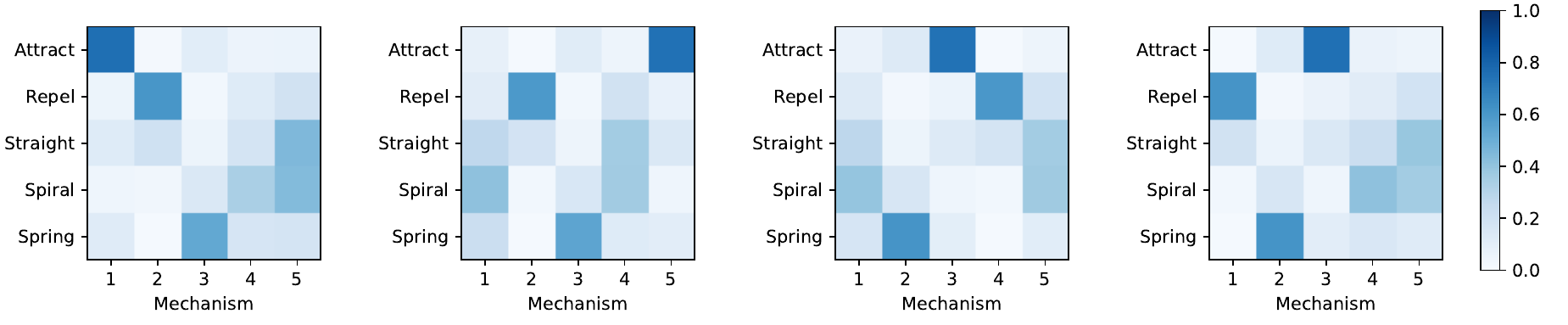}
    \caption{Disentanglement plots for different random seeds. We observe that similar structures emerge across different seeds (up to permutation). }
    \label{fig:random_seeds}
    \vspace{5mm}
\end{figure*}
\label{app:ablation_studies}
In practice, we find that the mechanism disentanglement of COMET is significantly improved with 1) warm starting the training process and 2) increasing the time horizon for mechanism selection. With these two modifications, we find that COMET can disentangle mechanisms robustly. In Fig. \ref{fig:random_seeds} we show that the emergence of the discrete mechanisms is robust across different random seeds. 

\subsection{Warm-starting Training}
One failure mode of COMET is that, due to the competitive nature of the training process, one or two mechanisms can dominate the competition and win all of the data. This can happen if the mechanisms are initialised randomly. In this case, one mechanism can, by chance, be the closest approximation to the environment dynamics and keep winning the competition. Fig. \ref{fig:no_warm_start} shows an example of this failure case, where only two mechanisms are used. In order to alleviate this, we add a warm-start phase before the actual competitive training, where the gradient is passed through all mechanism-context pairs regardless of accuracy. The intuition is that this allows all mechanisms to be roughly correct and as such have a fair chance of specialising in particular mode of interactions. 

\begin{figure*}[h]
    \centering
    \includegraphics[width=\textwidth]{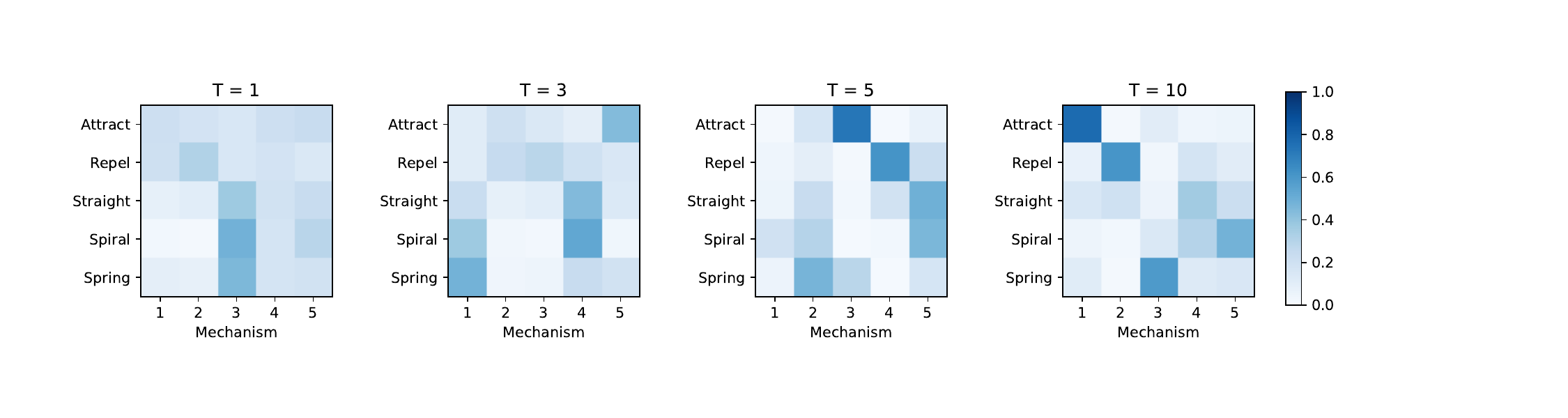}
    \caption{Effect of increasing the effective time horizon in the competition phase. As the time horizon increases, the quality of disentanglement improves.}
    \label{fig:time_horizon}
    \vspace{5mm}
\end{figure*}
\subsection{Increased Time-horizon}
Another failure mode that we encountered is that the model can learn non-interpretable mechanisms which leads to flickering mechansism selection during rollout. To this end, we found that increasing the effective time-horizon of the chosen mechanism-context pair is conducive to the emergence of well disentangled mechanisms. This can be implemented as a modification to the loss function, such that:
\begin{equation}
    \mathcal{L}(\theta_{0:M}) = \sum_{i=0}^K \min_{m,j} \left[\sum_{\tau=0}^T d\left(\mathbf{z}_i^{t+\tau} + \Delta\mathbf{z}^{t+\tau}_i(m,j), \mathbf{z}_i^{t+\tau+1}\right)\right],
\end{equation}
where $T$ is the time horizon which is set as a hyperparameter. 
Note that the sum over the time horizon is inside the minimisation, meaning that a mechanism-context pair is picked only if it has the best performance over multiple consecutive time steps. 
This serves as an inductive bias that encourages the sparsity of mechanism changes across time steps, which empirically results in better disentangled mechanisms. Fig. \ref{fig:time_horizon} shows the effect of increasing the time horizon. We observe that without a long time horizon, each mechanism learns a mixture of different modes of interactions. In our experiments, we set the time horizon to 10 for Particles, 3 for Traffic and 5 for Team Sports.

\section{Extra Rollouts}
\label{app:extra_rollouts}
\begin{figure*}[t]
    \centering
    \includegraphics[width=\textwidth]{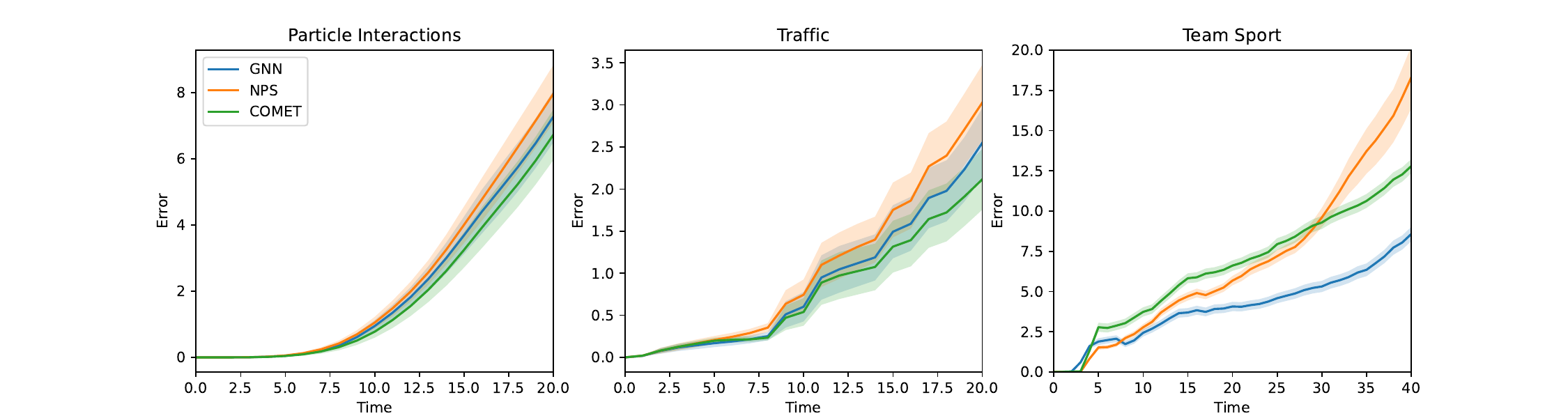}
    \caption{Given enough data, all model achieves commensurate rollout errors in the Particle Interactions and Traffic dataset. In the Team Sports dataset, COMET performs similarly to NPS, which also considers binary interactions, whereas GNN performs better. This suggests that only considering binary interactions is not sufficient in this environment. Despite this, COMET is still able to disentangle some interaction mechanisms.}
    \label{fig:extra_rollout}
    \vspace{5mm}
\end{figure*}
Fig. \ref{fig:extra_rollout} shows rollout plots for all of the models in a new environment in the large data limit. This shows that all model eventually converge to the same level of performance given enough data. Note that in the Team Sport environment is designed to be challenging for COMET as it requires reasoning about the entire scene, i.e. high order interactions between multiple objects. Here, GNN outperforms both NPS and COMET, which perform at a similar level. This supports our hypothesis that considering binary interactions can hinder model expressiveness. Nonetheless, we show that COMET is still able to disentangle mechanisms and adapt much faster than NPS, which also only considers binary interactions. 
\end{document}